%% file: ISWC2025-355.tex
\documentclass[runningheads]{llncs}

\usepackage[T1]{fontenc}
\usepackage{graphicx}
\usepackage{appendix}
\usepackage{todonotes}
\usepackage{multirow}
\usepackage[normalem]{ulem}

\usepackage{paralist}
\usepackage{enumitem}
\usepackage{comment}

\begin{document}
\title{RELRaE: LLM-Based Relationship Extraction, Labelling, Refinement, and Evaluation}
\titlerunning{LLM-Based RELRaE}
%
%

\newif\ifreview


\ifreview
    \author{Submission:   355}
    \institute{}
\else
\author{George Hannah\inst{1}\orcidID{0000-0002-3218-4559} \and
Jacopo de Berardinis\inst{1}\orcidID{0000-0001-6770-1969} \and
Terry R. Payne\inst{1}\orcidID{0000-0002-0106-8731} \and
Valentina Tamma\inst{1}\orcidID{0000-0002-1320-610X} \and
Andrew Mitchell\inst{2} \and
Ellen Piercy\inst{2} \and
Ewan Johnson\inst{2} \and
Andrew Ng\inst{2} \and
Harry Rostron\inst{2} \and
Boris Konev\inst{1}\orcidID{0000-0002-6507-0494}}
\authorrunning{G. Hannah et al.}
%
\institute{Department of Computer Science, University of Liverpool, Brownlow Hill, Liverpool, L69 7ZX, United Kingdom \and
Unilever Plc. Materials Innovation Factory, University of Liverpool, 51 Oxford Street, Liverpool, L7 3NY, United Kingdom \\
\email{g.t.hannah@liverpool.ac.uk}
\email{Jacopo.De-Berardinis@liverpool.ac.uk}
\email{trp@liverpool.ac.uk}
\email{valli@liverpool.ac.uk}
\email{Andrew.Mitchell@unilever.com}
\email{ellen.piercy@unilever.com}
\email{ewan.johnson@unilever.com}
\email{andrew.ng@unilever.com}
\email{harry.rostron@unilever.com}
\email{konev@liverpool.ac.uk}
}
\fi


\maketitle              
\begin{abstract}
A large volume of XML data is produced in experiments carried out by robots in laboratories. In order to support the interoperability of data between labs, there is a motivation to translate the XML data into a knowledge graph. A key stage of this process is the enrichment of the XML schema to lay the foundation of an ontology schema. To achieve this, we present the RELRaE framework, a framework that employs large language models in different stages to extract and accurately label the relationships implicitly present in the XML schema. We investigate the capability of LLMs to accurately generate these labels and then evaluate them. Our work demonstrates that LLMs can be effectively used to support the generation of relationship labels in the context of lab automation, and that they can play a valuable role within semi-automatic ontology generation frameworks more generally.

\end{abstract}
\input{Sections/1.Introduction}
\input{Sections/2.Background}
\input{Sections/3.Framework}
\input{Sections/4.PromptEngineering}

\input{Sections/5.Evaluation}

\section{Discussion}\label{sec:dis}

The results presented in Section~\ref{sec:exp} demonstrate that LLMs can bring value to the process of enriching an XML schema into a simple OWL ontology. If a lower similarity threshold is used, nearly all of the knowledge generated by an LLM can be injected into ontology. LLMs also show promise in their ability to evaluate this knowledge prior to its inclusion in the ontology. The \emph{Refined} method (based on the first three stages of the RELRaE pipeline)
was clearly the most accurate method across all test cases. This is likely due to fact that it combines the strengths of both the rules-based approach and the LLM only approach. Using the rules-based label (i.e. the RuBREx module) as a starting point allows a labelling protocol to be introduced by the ontology engineer, whilst the abstract reasoning capability of LLMs allows the label to be refined to a point where it accurately represents the true nature of the relationship. Another benefit of using the rules-based method as a starting point for refinement is that it limits the scope of the task. This effectively anchors the LLM to a simple way of interpreting the relationship such that the generated label, whilst more accurate, is not completely foreign to the original label.

Whilst the results presented in this work are promising and provide a valuable contribution in assessing the viability of LLMs in the ontology engineering process, the methods applied in this paper are not without limitation. A major limitation of this work is the way in which the ground truth was defined for this work, based only on a limited number of experts. A more consensual ground truth would improve the findings as the subjectivity inherent to the evaluation of the quality LLM responses would be reduced.
Additionally, in Section~\ref{sec:exp_eval}, when considering ``Possible'' evaluations, it is not certain that LLM is genuinely evaluating the label as ``Possible'', or whether the label is not clearly incorrect enough to be considered in the ``Unlikely'' or ``No'' categories. A potential solution to this limitation would be to prompt the LLM to generate a ``ground truth'' (or alternate) label for the relationship and then computing the cosine similarity between the two labels. However, due to the subjective nature of evaluating labels, it is impossible to have perfect evaluation.

There is a need for future work to expand on the findings of this paper. One possible direction would be a comparative evaluation of different LLMs for each task. In this work, we assume that there is little difference in the quality of labels generated or their evaluation depending on which LLM is used; however other studies have demonstrated that the performance of different LLMs can vary, depending on the task and the benchmark used~\cite{white2024livebench}.

Whilst our results strongly suggest that using an LLM in this knowledge engineering task is worthwhile, it is important to note that the performance of the naïve rules-based method may also be considered acceptable.
Nevertheless, RELRaE's approach extends the former while still keeping its benefits, including the consistency of the generated labels and the additional control of this method.

\section{Conclusion}\label{sec:conc}
In this work we present the RELRaE framework as a method to translate an XML schema into a simple ontology by making explicit the implicit semantics present in the schema. This is done by extracting all of the hierarchical relationships present in the schema. Labels are then generated for these relationships. We investigate the method by which these labels are generated, comparing three alternate methods. We conclude that whilst all methods can produce acceptable labels, a hybrid approach using a rules set to generate a simple label for the relationship first and then using an LLM to further refine this label leads to the highest quality labels. Additionally, we assess the viability of LLMs as evaluators of these generated labels, in an attempt to mitigate an potential hallucinations the LLM may generate. We find evidence to suggest that an LLM can successfully evaluate the generated labels, however the quality of each evaluation is still unknown. This work provides a valuable contribution in evidence that suggests that LLMs can provide value in the ontology engineering process, specifically in the process of enriching an XML schema into an ontology.

~\newline

\noindent{\bf Supplemental Material Statement:} For the purposes of reproducibility, all code, datasets and evaluation tools are available on GitHub.\footnote{\url{https://anonymous.4open.science/r/RELRaE}}

\begin{credits}
\subsubsection{\ackname}  
\ifreview
The first author
\else
George Hannah
\fi
has been funded by an EPSRC ICASE studentship
\ifreview
with an Industrial Partner (redacted for double blind reviewing purposes).
\else
(201146) with Unilever PLC.
\fi

\subsubsection{\discintname}
\end{credits}
\bibliographystyle{abbrv}
\bibliography{bibliography}



\end{document}

%% file: Sections/1.Introduction.tex
\section{Introduction} \label{sec:intro}

Semi-structured data formats, including eXtensible Markup Language (XML) and Comma Separated Values (CSV), represent a vast amount of data on the web today \cite{W3C_CSVW_UCR}.
These formats typically offer implicit semantics, connecting concepts within the documents.
However, structured data formats like the Resource Description Framework (RDF) provide explicit, machine-readable semantics where relationships between concepts are clearly articulated.
XML is usually defined by an XML Schema Document (XSD), whereas ontologies are used to define Knowledge Graphs (KGs).
These ontologies (typically expressed in RDF Schema (RDFS) or the Web Ontology Language (OWL)) define concepts and their interrelations that allow inference and offer a richer view of the data within KGs~\cite{tiddi2022knowledge}.


Harnessing explicit semantics is crucial in data-intensive science.
For example, analytical chemistry labs using robots generate vast quantities of data at an unprecedented rate over manual methods \cite{tom2024self}.
This data is often captured in XML-based formats like the Analytical Information Markup Language\footnote{AnIML core schema: \url{https://www.animl.org/current-schema}} (AnIML) or the Allotrope Simple Model\footnote{\url{https://www.allotrope.org/asm}} (ASM).
While their XSDs (e.g., in AnIML) define structure (like \emph{Parent-Child} or \emph{Element-Attribute} links), the deeper meaning of scientific concepts frequently remains semantically underspecified.
This semantic gap presents a challenge: elevating such implicitly structured scientific data to an explicit, knowledge-rich representation.
Bridging this semantic gap involves translating XSDs into ontologies (e.g., OWL or RDFS).
Existing methods \cite{hacherouf2015transforming} often use rule-based exploitation of XML's hierarchical structure~\cite{bedini2011transforming,bohring2005mapping,hajjamy2017xsd2owl2,zhang2022constructing}.
Whilst these approaches offer a starting point in the translation of XSD into OWL, the inherent KG-XML difference (explicitly structured vs. semi-structured) means effective XSD-to-ontology translation requires extensive knowledge and expertise in these domains.
In traditional ontology engineering methods, this is done by a domain expert (DE)~\cite{noy2001ontology,sequeda2017pay}, a process that can be time-consuming and prone to knowledge acquisition bottlenecks  \cite{feigenbaum1977art,neches1991enabling}.

In this study, we aim to address the limitations of previous approaches by integrating Large Language Models (LLMs) into the XSD-to-ontology translation process, with the additional aim of reducing the workload of the DE and the ontology engineer.
We present \textbf{RELRaE} (Relationship Extraction, Labelling, Refinement, and Evaluation), a novel framework for translating an XML schema into a foundational RDFS ontology.
The basic ontology produced by RELRaE can be described as a \textit{skeleton ontology}, representing the inter-concept relationships present in the XML schema input with injected domain knowledge.
RELRaE operates in four distinct stages: (i) structural information representing hierarchical relationships between candidate concept pairs is extracted from the XML schema (\textit{Concept Relationship Extraction}); (ii) labels for these relationships are then generated using a rule-based approach (\textit{Rule-based Label Generation}); (iii)  an LLM then refines these initial labels using schema-based contextual information (\textit{Label Refinement}); and finally (iv) a different LLM, acting as a proxy for a domain expert, assesses the suitability of the refined label (\textit{Automatic Label Evaluation}).
This multi-stage process aims to produce a robust skeleton ontology, providing a strong starting point for further enrichment activities such as taxonomy completion~\cite{shi2024taxonomy} and ontology alignment~\cite{jimenez2011logmap}.

To validate our approach, we conducted a comparative empirical evaluation using the AnIML schema, providing a complex benchmark from the analytical chemistry domain.
This compared relationship labels generated by RELRaE's hybrid approach against those from purely rule-based and LLM-only methods, using a gold-standard reference set created by domain experts.
We also investigated the efficacy of an LLM in evaluating the generated labels by comparing its assessments against those of a domain expert.
The findings indicate that RELRaE significantly enhances label accuracy compared to other methods and that LLMs demonstrate promise for the automated evaluation of these labels.

The contributions of this paper are twofold: (i) RELRaE, a structured, multi-stage method that synergistically combines rule-based techniques with LLMs for the generation of skeleton ontologies from XML schemata; (ii) an empirical evaluation of RELRaE, focusing on its ability to generate accurate relationship labels and the viability of the LLM-as-a-judge approach for label assessment within a challenging, domain-intensive benchmark.
Overall, this work demonstrates that the reuse of LLMs within a structured framework like RELRaE, can effectively support semi-automatic ontology generation, particularly in a complex domain.


%% file: Sections/2.Background.tex
\section{Background and Related Work}\label{sec:lit}

Several studies
have explored the notion of enriching XSD to generate an OWL ontology, with many approaches being rule-based \cite{hacherouf2015transforming}.
This relies on the assumption that the semantics present in the hierarchical structure of XML are implicit. 
Bohring and Auer \cite{bohring2005mapping} proposed one approach to translate XML schema directly into an OWL ontology via a mapping based pipeline, by identifying the structural patterns that define the elements and their relationships in the XML schema, and mapping them to a corresponding OWL pattern. Each relationship identified in the source schema is classed as either a \emph{partOf} or a \emph{subClassOf} relation, and labels for these relationships are formed by attaching either the ``\emph{has}'' or ``\emph{dtp}'' prefix to the textual label of the entity.
Whilst their rules can perform well in situations where the source XSD is well formed and where standard structural patterns are used to represent the relationships between concepts, this is not always the case, and the relationships identified do not necessarily correlate with the intentions of the schema author.
These limitations can be addressed through the \emph{injection} of additional knowledge, thereby enriching the knowledge inherent in the XSD to construct a new ontology.  Within traditional ontology engineering methodologies this knowledge is typically provided by the domain expert~\cite{noy2001ontology}; however, this process can be significantly time consuming and suffers from the problem of the knowledge acquisition bottleneck.


The rise in popularity of LLMs since the release of ChatGPT-3 in late 2022\footnote{\url{https://openai.com/index/chatgpt/}} has resulted in a number of investigations into the viability of utilising LLMs as part of the ontology engineering process~\cite{Alharbi2024_ESWC,alharbiEKAW2024,he2023exploring,Hertling_2023,lippolisontogenia,qiang2024agentomleveragingllmagents,saeedizade2024navigating}. Due to the large volume of training data used in their creation, LLMs have a large breadth of knowledge across many domains, which can be exploited as a source of additional (and in many cases domain specific) knowledge, thereby supplanting the domain expert, at least with respect to modelling the relationships between more general concepts. Thus, LLMs can potentially play a valuable role as part of an ontology engineering pipeline, as they reduce the dependency on the domain expert, allowing them to focus on the fine-grain nuanced knowledge, crucial for accurately representing the domain~\cite{song2025injecting}.
Additionally, LLMs possess strong natural language processing (NLP) ability, allowing them to parse and extract information from large, unformatted bodies of text. This ability is valuable to the ontology engineering process as they can be employed to extract vast amounts of domain specific knowledge from domains that are well represented in scientific literature, and incorporate them within an ontology~\cite{babaei2023llms4ol}.


A key aspect regarding the utilisation of LLMs is the format, or construction of the prompt used. The presence of certain prompt features can significantly augment the quality of the generated response. Prompt-engineering approaches aim to construct prompts that return the highest possible quality response from the LLM for a given problem or task~\cite{hannah2025legal}. A number of different principles have been defined to attempt to isolate the specific syntactic features that cause the LLMs to modify their responses~\cite{bsharat2023principled}, however due to the black box nature of LLMs, the reasons why such modifications in the response occur is still unclear.

One disadvantage of LLMs in these situations is that LLMs are prone to hallucinations, i.e. the generation of false information~\cite{ji2023survey}. 
To ensure that these hallucinations are identified early in the ontology engineering process, the quality of the generated knowledge needs to be evaluated. 
However, this comes at a cost, due to the fact that as the size of the ontology expands, the time required by DEs to evaluate the LLM-generated knowledge also increases. To mitigate this, the use of the LLM-as-a-judge approaches have emerged~\cite{barile2025lp,desmond2024evalullm,shankar2024validates}, whereby an LLM is also used to evaluate the suitability, or correctness of a separate LLM-generated response with respect to the domain of a specific use-case.
Whilst these approaches show promising performance, limitations are also present in these methods with regards to the LLM's understanding of more complex, domain specific knowledge~\cite{li2024llms,szymanski2025limitations}, and thus the knowledge of a domain expert is still critical to the creation of a high quality representation of their specific domain.


%% file: Sections/3.Framework.tex
\section{RELRaE Framework}\label{sec:fram}


The organisation and structure of data within specialised data-rich domains is typically defined through community or commercially defined schemata, which are typically annotated and documented.  However, in many domains, these schemata can be semantically-incomplete, as there is a focus on the organisation and labelling of content with implicit relationships between data elements.
To facilitate broader reuse, such as increasing interpretability across different commercial organisations to support collaboration, or with other standards, the schemata need to be augmented with 
knowledge-rich definitions of those relationships through, for example, the use of an ontology.
However, enriching schemata with additional domain-specific knowledge is non-trivial, requiring significant engagement with domain experts. In this section, we  present \emph{RELRaE}, a novel framework that proposes and refines knowledge-rich definitions of relationships between concepts through a hybrid use of rules and LLMs that play different roles throughout the pipeline.

The \emph{Relationship Extraction, Labelling, Refinement, and Evaluation} (RELRaE) framework constructs relationship labels by following four distinct stages:
\begin{description}[style=unboxed]
    \item [1: Concept Relationship Extraction.]
    Structural information representing the relationships between candidate concept pairs are extracted from the XML schema, based on their hierarchical configuration.
    \item [2: Rule-based Label Generation.]
    Using the elicited structural information, a rule-based approach generates possible labels for each of the relationships.
    \item [3: Label Refinement.]
    An LLM is prompted to determine if the label is acceptable, or could be refined, based on schema-based contextual information.
    \item [4: Automatic Label Evaluation.]
    The resulting label is then evaluated to determine its suitability given other domain knowledge, using a \emph{different} LLM acting as a proxy for a domain expert.
\end{description}
The framework processes all of the possible relationships in an XML schema to generate simple labels for these relationships. Once labels have been identified for the relationships identified for each of the concept pairs, they are used to generate simple predicate axioms, which are then merged to construct an ontological representation (i.e. a \emph{skeleton ontology}) of the original XML Schema.

The first two stages have been combined into the \emph{Rules-Based Relationship Extraction (RuBREx)} module. Different LLMs are used to evaluate the suitability (stage 3) and acceptability (stage 4) of the labels, with the acceptability being ranked using a 5-point Likert scale (Table \ref{tab:enum}).

\begin{table}[t]
    \caption{Relationship labelling patterns. Each row represents one pattern, with <d> and <r> used to denote the labels of the domain and range elements respectively. For patterns 5 and 6, <r> and <d> are stated to indicate that for these patterns the domain and range are reversed compared to the schema.}
    \setlength{\tabcolsep}{4pt} 
    \centering
    \scriptsize
    \begin{tabular}{c c c c c c}
         \hline
         \multirow{2}{*}{\bf ID} & {\bf Domain} & {\bf Intermediate} & {\bf Range} & \multirow{2}{*}{\bf Restriction} & {\bf Label} \\
         & {\bf Type} & {\bf Elements} & {\bf Type} &  & {\bf Template} \\
         \hline
         1 & ComplexType & Sequence & ComplexType & & {\sf has<r>} \\
         2 & ComplexType & Sequence & SimpleType & & {\sf has<r>} \\
         3 & ComplexType & Sequence & SimpleType & {\sf type=xsd:boolean} & {\sf is<r>} \\
         4 & ComplexType & Sequence, Choice & ComplexType & & {\sf <r> subclassOf <d>}\\
         5 & ComplexType & Sequence, Choice & SimpleType & & {\sf <r> subclassOf <d>}\\
         6 & ComplexType & & Attribute & & {\sf has<r>} \\
         7 & ComplexType & & Attribute & {\sf type=xsd:boolean} & {\sf is<r>} \\
         9 & SimpleType & & & & {\sf Datatype} \\
         10 & Attribute & & & & {\sf Datatype} \\
         \hline
    \end{tabular}
    \label{tab:rel_patts}
\end{table}

\subsection{Stage 1\&2: Rules-Based Relationship Extraction (RuBREx)}

This module is responsible for the first two stages of the pipeline. Given an XML schema file, the module attempts to identify all of the relationships linking two concepts, where the concepts correspond to different entities (including elements, attributes, constraints etc.) in the schema.  A \emph{relationship label} is defined for pairs of entities where the entities have an implicit hierarchical relationship; which can be \emph{direct} (such as a parent-child relationship), or \emph{indirect} due to the existence of intermediate elements, such as a grandparent-grandchild relationship, where some parent element is found in the hierarchy between the other two entities. The type of element affects its characterisation in the class, following the approach suggested by Bohring \& Auer~\cite{bohring2005mapping}; {\sf complexType} elements (i.e. XML elements that contain other elements and/or attributes) are represented as {\sf owl:class} resources, whereas {\sf simpleType} elements (which specify the constraints and information about the values of attributes or text-only elements) are extracted as they provide additional contextual information and may also appear in the ontology (though not as classes). Furthermore, any documentation associated with the two entities, such as that included in an {\sf annotation} element, is also extracted, as it may be used in the stage 3, the label refinement stage.



Once the relationships have been extracted from the schema, a set of structural patterns (Table~\ref{tab:rel_patts}) are used to construct possible labels that relate the \emph{domain} concept to the  \emph{range} concepts (these correspond to class definitions of the concepts extracted from the schema), based on~\cite{bohring2005mapping}.  If no pattern is used, the approach defaults to generating a simple label (similar to patterns 1 and 2), by applying the prefix ``has'' to the label of the 
\ifreview
range concept.
\else
range concept~\cite{hannah2023towards}.
\fi
This results in a triple of the form {\sf [Domain, Relationship, Range]}.

\subsection{Stage 3: Label Refinement}

This third phase is responsible for querying an LLM in order to determine if the label emerging from the first two stages (i.e. the RuBREx module) could be refined, given the context in which it is used.  This requires the development of a prompt that provides the context of a domain, which is constructed from several textual fragments.  Initially, the notion that there is a requirement for an ontological label for a relationship should be specified:
\begin{description}[rightmargin=0.5cm,topsep=2pt]
    \item[~]
        \emph{``...this relationship has a simple label describing it and will be used in an ontology representing the domain of ...''}
\end{description}
\noindent
This context, together with the label and the two concepts form part of a prompt (discussed in Section \ref{sec:prompt_1}).  Specifically, the prompt asks whether the label presented is appropriate for the relationship given the domain and range concepts:

\begin{description}[rightmargin=0.5cm,topsep=2pt]
    \item[~]
        \emph{``...Does this label accurately represent this relationship?...''}
\end{description}
\noindent
However, if the label is not considered appropriate, then the LLM is asked to propose a better label:
\begin{description}[rightmargin=0.5cm,topsep=2pt]
    \item[~]
        \emph{``...If it doesn't, refine the label to more accurately represent the relationship...''}

\end{description}

In order to further refine the result, the prompt uses a few-shot strategy, whereby examples of the type of labels expected are given (see Tables \ref{tab:examples_1} and \ref{tab:examples_2} for examples of generic and domain specific examples, respectively).  The advantage of using this approach is that it provides the LLM with exemplar responses for a set of relationships that cab improve the quality of the resulting responses~\cite{hannah2025legal}. The full details of the prompt construction are discussed in Section~\ref{sec:prompt_1}.




\subsection{Stage 4: Automatic Label Evaluation} \label{sec:label_eval}

Ideally, when creating the labels for different concepts in domains that contain specialised knowledge (such as the autonomous analytical chemistry), a domain expert should be employed to assess the quality of the resulting labels, to verify their veracity.  However, this process requires significant levels of human effort and expertise. In this stage, a separate LLM is employed to act as an evaluation or proxy agent (thus reducing the dependency of using domain experts) to perform the label assessment. This follows similar work to that exploring the notion of an \emph{LLM-as-a-judge}~\cite{barile2025lp,desmond2024evalullm,shankar2024validates}, whereby an LLM is also used to evaluate the suitability, or correctness of a separate LLM-generated response.  To reduce any possible bias in the assessment, an LLM from a different model family is used to provide an assessment using a five-point Likert scale, as illustrated in Table~\ref{tab:enum}, thereby promoting consistency in the LLM-based assessment of labels. 
All refined labels that fall into either the ``Likely'' or ``Yes'' categories are accepted;\footnote{The choice to use either ``Likely'' or ``Yes'' is based on a preliminary study, and is reinforced through the results of the evaluation in Section \ref{sec:exp_eval}.} otherwise the LLM-based label from stage 3 is discarded and replaced with the original RuBREx label generated in stage 2.


\begin{table}[t]
    \centering
        \setlength{\tabcolsep}{8pt} 

    \caption{Mapping the score ranges with their associated (Likert scale) responses.}
    \begin{tabular}{c c c}
        \hline
         {\bf Score Range} & {\bf Likert Score} & {\bf Response} \\
         \hline
         0\% - 20\% & 1 & No \\
         20\% - 40\% & 2 & Unlikely \\
         40\% - 60\% & 3 & Possible \\
         60\% - 80\% & 4 & Likely \\
         80\% - 100\% & 5 & Yes \\
         \hline
    \end{tabular}
    \label{tab:enum}
\end{table}


Once all of the relationships identified for concept pairs within the original XML schema have been identified, they are each translated into 
a corresponding RDF fragment. This fragment is then merged to form an initial, or \emph{skeleton} ontology representing the identified concepts and meaningful labels.  This ontology can then be further refined using some other ontology enrichment processes, such as using taxonomy completion~\cite{shi2024taxonomy}, or alignment to other ontologies~\cite{faria2013agreementmakerlight,jimenez2011logmap}.

%% file: Sections/4.PromptEngineering.tex
\section{Prompt Engineering}\label{sec:prompt}

The design of an LLM prompt can greatly affect the quality and utility of the response \cite{Brown2020fewshot,LiuPrompt_2023,liu2024jailbreaking,Marvin2024promt}, including the definition of context, clarity over the question being asked and additional training examples through few-shot queries. Therefore, the evolution of a usable prompt can involve iterating over a set of refinements, starting with an embryonic or baseline prompt with a preliminary set of examples.  The RELRaE framework employs two separate LLMs: LLM$_{\mathcal{R}}$ which is used in stage 3 (label refinement), whereby a relationship label should be modified given additional contextual information; and LLM$_{\mathcal{E}}$ which is used in stage 4 (label evaluation), whereby the LLM acts as a proxy for a domain expert and assesses the validity of the label given the domain. Each LLM has its own prompt that has been developed for its specific goals, discussed below. In this section, we discuss the candidate prompts to illustrate this iterative process, and then characterise the final prompts that are used in the evaluation in Section \ref{sec:exp}.
In order to contextualise this within a specific domain, we focus on prompts that are tailored to the analytical chemistry domain, and evaluate the approach using the 
Analytical Information Markup Language\footnote{AnIML core schema: \url{https://www.animl.org/current-schema}} (AnIML) schema.


\subsection{Label Refinement Prompt} \label{sec:prompt_1}


During stage 3 (label refinement), LLM$_{\mathcal{R}}$ has two subtasks: i) to determine whether or not the candidate RuBREx label (generated by the first two stages) is appropriate for a given relationship between two concepts, and if so to return that label; and ii) to refine the label if the candidate label was not considered appropriate, and return the resulting (refined) label.


An initial, or \emph{base} prompt was initially generated  containing the core instructions for the label-refinement LLM:
\begin{description}[rightmargin=0.5cm,topsep=2pt]
    \item[Base prompt:]
        \emph{``Given a relationship between two concepts, this relationship has a simple label describing it. Does this label accurately represent this relationship? If it doesn't, refine the label to more accurately represent the relationship. If it does, return the original label.''}
\end{description}
\noindent
The base prompt lacks sufficient domain specific context to retrieve meaningful labels.  Therefore, as in this study it will be applied to
translating AnIML into an ontology, and in noting the domain itself (analytical chemistry). Thus, the final refinement prompt is defined (the contextual extensions are shown in bold):
\begin{description}[rightmargin=0.5cm,topsep=2pt]
    \item[Refinement prompt:]
        \emph{``\textbf{You are an analytical chemist.} Given a relationship between two concepts \textbf{linked in an XML schema}, this relationship has a simple label describing it \textbf{and will be used in an ontology representing the domain of chemical formulation experiments being carried out by robots in a lab}. Does this label accurately represent this relationship? If it doesn't, refine the label to more accurately represent the relationship. If it does return the original label.''}
\end{description}
\noindent


An extended prompt was also considered whereby additional relation-specific context was provided. In this case, the prompt included any documentation found within the schema that was pertinent to the concepts associated with the relationship. The format of this extended prompt is given below (again the contextual extensions to the refinement prompt are shown in bold):
\begin{description}[rightmargin=0.5cm,topsep=2pt,style=unboxed]
    \item[Refinement with documentation prompt:]
        \emph{You are an analytical chemist. Given a relationship between two concepts linked in an XML schema \textbf{and defined with the following documentation}, this relationship has a simple label describing it and will be used in an ontology representing the domain of chemical formulation experiments being carried out by robots in a lab. Does this label accurately represent this relationship? If it doesn't, refine the label to more accurately represent the relationship. If it does return the original label}.
\end{description}
\noindent

An approach for targetting or training an LLM to provide more specific results is through the use of \emph{few-shot} prompts~\cite{liu2024jailbreaking}, whereby positive examples can be included in the prompt. 
Two scenarios were considered: i) generic examples (Table~\ref{tab:examples_1}) that illustrate improved relationship names for arbitrary concepts; and ii) domain-specific examples (Table~\ref{tab:examples_2}) that illustrate examples that are much more relevant to the task. The intuition here was that although domain-specific examples had more relevance to the specific task, there was the possibility of being too specific and thus biassing the final results to reflect the examples.




\begin{table}[t]
    \caption{Generic Relationship Examples.}
    \centering
    \setlength{\tabcolsep}{12pt} 

    \begin{tabular}{l l l l}
        \hline
        {\bf Domain} & {\bf Range} & {\bf RuBREx Label} & {\bf LLM Response} \\
        \hline
        Apple & Fruit & {\sf subClassOf} & {\sf typeOf} \\
        Business & Cashier & {\sf hasCashier} & {\sf hasPosition} \\
        Fingers & Hand & {\sf subClassOf} & {\sf partOf} \\
        Keyboard & Input Device & {\sf subClassOf} & {\sf subClassOf} \\
        Father & Son & {\sf hasSon} & {\sf hasChild} \\
        \hline
    \end{tabular}
    \label{tab:examples_1}
\end{table}

\begin{table}[t]
    \caption{Domain-specific Relationship Examples.}
    \centering
    \setlength{\tabcolsep}{12pt} 

    \begin{tabular}{l l l l}
        \hline
        {\bf Domain} & {\bf Range} & {\bf RuBREx Label} & {\bf LLM Response} \\
        \hline
        Argon & Nobel Gases & {\sf hasNobelGases} & {\sf memberOf} \\
        Template & template ID & {\sf hasTemplateID} & {\sf hasID} \\
        Tag Set & Tag & {\sf hasTag} & {\sf hasMember} \\
        Author & Role & {\sf hasRole} & {\sf hasRole} \\
        Technique & URI & {\sf hasURI} & {\sf hasURI} \\
        \hline
    \end{tabular}
    
    \label{tab:examples_2}
\end{table}

\noindent
{\bf Prompt Evaluation:}
The above prompts were evaluated across a small number of examples taken from the AnIML schema, to determine which was likely to result in the best performance before being employed by the full RELRaE framework.
OpenAI's GPT-4 model\footnote{\url{https://openai.com/research/gpt-4}} was selected as LLM$_{\mathcal{R}}$
This evaluation found that the prompt providing a role as well as the source domain and the final use-case of the relationship, provides the highest quality responses from the prompts that were tested (i.e. the \emph{Refinement Prompt}).
This type of prompt offered a good balance between context and conciseness. 

When using the \emph{Base Prompt}, for relationships involving more generalisable concepts, the LLM$_{\mathcal{R}}$ often struggles to understand the specific domain of AnIML, but benefits from the inclusion of the schema documentation.
Whilst in many cases this would be a desirable behaviour, for  the AnIML schema, there is a lack of consistency in the documentation which compromised the overall performance. For example, some concepts were defined several times in the schema yet possessed inconsistent documentation for each definition.
Furthermore, when comparing the few-shot approach with generic- and domain-specific examples, little difference in performance was observed. However,  the response quality for both few-shot scenarios was found to be higher than the zero-shot counterparts.


Thus, for the purposes of the main evaluation, a few-shot variant of the \emph{Refinement Prompt} was selected with domain-specific examples; based on the assumption that these examples would guide the LLM$_{\mathcal{R}}$ in correctly interpreting the complex domain of analytical chemistry for the more technical relationships. 


\subsection{Label Evaluation Prompt} \label{sec:prompt_2}

\begin{table}[t]
    \centering
    \setlength{\tabcolsep}{8pt} 

    \caption{Examples evaluations provided to the LLM.}
    \begin{tabular}{l l l l}
         \hline
         {\bf Domain Label} & {\bf Range Label} & {\bf Relationship Label} & {\bf Likert Score} \\
         \hline
         Sample & Barcode & {\sf hasBarcode} & Yes \\
         derived & xsd:boolean & {\sf isDerivedSample} & No \\
         Method & name & {\sf hasOptionalMethodName} & Likely \\
         Method & Author & {\sf wasPerfomedBy} & Possible \\
         id & xsd:ID & {\sf UniqueIdentifier} & Unlikely \\
         \hline
    \end{tabular}
    
    \label{tab:examples_3}
\end{table}
The aim of the LLM$_{\mathcal{E}}$ in stage 4 (label evaluation) is to determine if the label constructed in stage 3 is suitable for the domain, though the use of a Likert scale (Table~\ref{tab:enum}).  This is used to facilitate the LLM$_{\mathcal{E}}$ in generating 
a valid evaluation of the relationship label.
%
An iterative approach was used to identify and refine the prompt for this task, by injecting increasing levels of context into the prompt. An initial, context-sparse prompt was constructed to determine if the label given accurately represents the relationship between two concepts. This was extended by adding the definition of a relationship, derived from the definition of a property in the RDF specification~\cite{w3ResourceDescription}, and 
adapted for the final iteration of the prompt, which included any documentation present in the schema for each concept, thus providing the LLM$_{\mathcal{E}}$ the context in which the concepts are interacting:




\begin{description}[rightmargin=0.5cm,topsep=2pt,style=unboxed]    \item[Evaluation Prompt]
        \emph{``Is the label \textsf{label} an appropriate label to describe the relationship between domain: \textsf{domain} and range: \textsf{range}? These concepts are derived from an XML schema describing the domain of chemical formulation experiments carried out by robots in laboratories. The relationship linking these concepts will be used in an ontology representing the same domain.''}
\end{description}

\noindent
{\bf Prompt Evaluation:}
The above prompts were evaluated across a small number of examples based on the AnIML schema. Gemini-2.0-flash~\cite{team2023gemini} was chosen for LLM$_{\mathcal{E}}$, as its API allows for structured outputs in the format of python enums. This allows us to constrain the LLM$_{\mathcal{E}}$'s responses to those used by the five-point Likert scale. A few-shot approach was also used, by including the the examples listed in Table~\ref{tab:examples_3} with the prompt.
The results from the preliminary evaluation suggest that using a concise prompt, with a small amount of context and a set of examples at a low temperature setting (i.e. 0.3) 
generated acceptable evaluations.

%% file: Sections/5.Evaluation.tex
\section{Methodology and Evaluation}\label{sec:exp}

The premise behind RELRaE is that LLMs can be used to refine simpler, heuristic approach to label generation (such as the rule based approach used by RuBREx), thus generating higher quality (i.e. more relevant) labels for the relationships between concepts.  Furthermore, other LLMs can also act as a proxy to a domain expert in assessing the suitability of the resulting labels for a given domain. 
This resulted in the following two research questions: 

\begin{description}[topsep=2pt,style=unboxed]
    \item[RQ1: ]
        To what extent can an LLM capture the nature of a relationship between two concepts and produce an acceptable label from this relationship?
    \item [RQ2:]
        How close is an LLM's ability to evaluate a relationship's label to a domain expert's ability to evaluate that same label?
\end{description}

\noindent
Both questions are investigated empirically in this section.  Section \ref{sec:exp_ref} presents a comparative analysis of three alternate approaches for generating the labels of the relationships between two concepts.  A reference set of labels has been constructed by three domain experts, and used to evaluate the suitability of the automatically generated labels, thereby addressing RQ1.  The second evaluation (Section \ref{sec:exp_eval}) assesses the efficacy of using an LLM as a proxy for a Domain Expert, thereby addressing RQ2.  This was conducted by comparing the assessment of the suitability of labels from an LLM with that of a Domain Expert, through a five-point Likert scale.



\subsection{Label Refinement with LLM$_{\mathcal{R}}$}\label{sec:exp_ref}

To understand the efficacy of using an LLM to refine labels (i.e. LLM$_{\mathcal{R}}$), a comparative analysis of three label generation methods was performed:
\begin{description}[topsep=2pt,style=unboxed]
    \item[RuBREx Labels:] A rule-based only approach was used to generate labels for relationships.  This corresponds to stages 1\&2 of the RELRaE pipeline, and utilises the rules given in Table \ref{tab:rel_patts}.

    \item[Refined Labels:] This utilises the first three stages of the RELRaE pipeline to generate labels, and then refine them utilising an LLM$_{\mathcal{R}}$ using the \emph{Refinement Prompt} (Section \ref{sec:prompt_1}).

    \item[LLM-only Labels:] This baseline approach generates labels by querying the LLM$_{\mathcal{R}}$ directly (rather than asking an LLM to refine an existing label).  It makes use of the \emph{LLM-only Prompt}, described below. 
\end{description}

The main difference between the third approach (using the \emph{LLM-only Labels} prompt) and the use of \emph{Refined Labels} is that it explicit requests the generation of a new label (highlighted in bold below), but is not provided with an initial label.  The inclusion of details regarding the concepts, and the use of few-shot examples is consistent across both approaches.

\begin{description}[rightmargin=0.5cm,topsep=2pt,style=unboxed]

    \item[LLM-only Labels prompt:]
        \emph{``You are an analytical chemist. Given a relationship between two concepts linked in an XML schema, this relationship will be used in an ontology representing the domain of chemical formulation experiments being carried out by robots in a lab. \textbf{Generate a label to accurately represent this relationship.}''}        
\end{description}
\noindent

OpenAI's GPT-4o model\footnote{\url{https://openai.com/research/gpt-4}} was selected as the LLM$_{\mathcal{R}}$ to generate or refine the relationship labels. As the \emph{temperature} hyperparameter controls the ``creativity'' or stochasticity of the LLM$_{\mathcal{R}}$~\cite{ji2023survey}, the evaluations were performed using five temperature settings, $temp \in \{0.0, 0.25, 0.5, 0.75, 1.0 \}$.  These were selected to explore the hypothesis that as the temperature increases, the LLM's ability to capture the abstract nature of the relationships would also increase, whilst sacrificing the consistency of the responses (which is expected to decrease).

An expert-led set of labels were independently generated (by three domain experts) for all of the relationships that had been extracted from the AnIML schema; resulting in a gold-standard reference set of labels.  The performance of each approach was calculated by determining cosine similarity between the generated labels and those in the reference set.

In order to evaluate the performance of each of these approaches, Phrase-BERT~\cite{phrasebertwang2021} is used to calculate the cosine similarity between the generated label and the reference-set labels; the resulting scores are then used to determine a final \emph{acceptance} set of labels, based on a similarity threshold $\tau$.  Two thresholds are used to determine which labels were included in the final set: loose ($\tau \ge 0.6$) and strict ($\tau \ge 0.85$). These thresholds were determined empirically prior to the main evaluation using a small sample set of the generated labels. It should be noted, however, that by adopting this approach, if there is a significant lexical difference between the label (being evaluated) and the reference labels, then the similarity score could fall below the acceptance threshold, even if the generated label satisfactorily represents the nature of the relationship.  Conversely, there is the danger that labels that are  similar lexically could score highly, even if they fail to satisfactorily represent the relationship.


\begin{table}[t]
    \centering
    \setlength{\tabcolsep}{12pt} 

    \caption{Mean cosine similarity between the generated relationship labels and those in the reference set (using Phrase-BERT embeddings).}
    \begin{tabular}{c || c | c | c}
        $temp$ & RuBREx & LLM-only & Refined \\
        \hline
        0.00 & \multirow{2}{*}{$\vdots$} & 0.810 & {\bf 0.883} \\
        0.25 & & 0.808 & {\bf 0.880} \\
        0.50 & 0.809 & 0.814 & {\bf 0.880} \\
        0.75 & \multirow{2}{*}{$\vdots$} & 0.786 & {\bf 0.874} \\
        1.00 & & 0.793 & {\bf 0.861} \\
        \hline
    \end{tabular}
    \label{tab:refinement_exp_1}
\end{table}

\begin{table}[t]
    \centering
    \setlength{\tabcolsep}{10pt} 

    \caption{Percentage of labels who's cosine similarity score exceeded the acceptability threshold for two thresholds; loose similarity ($\ge 0.6$) and strict similarity ($\ge 0.85$).}
    \begin{tabular}{c || c | c | c | c | c | c}
        \multirow{2}{*}{$temp$} & \multicolumn{2}{c|}{RuBREx (\%)} & \multicolumn{2}{c|}{LLM-only (\%)} & \multicolumn{2}{c}{Refined (\%)} \\
        \cline{2-7}
        & $\ge 0.6$ & $\ge 0.85$ & $\ge 0.6$ & $\ge 0.85$ & $\ge 0.6$ & $\ge 0.85$ \\
        \hline
        0.00 & \multirow{2}{*}{$\vdots$} & \multirow{2}{*}{$\vdots$} & 90.6 & 38.8 & \textbf{96.4} & \textbf{57.6} \\
        0.25 & & & 91.0 & 36.7 & \textbf{96.4} & \textbf{55.8} \\
        0.50 & 86.3 & 49.3 & 92.8 & 38.5 & \textbf{96.0} & \textbf{56.8} \\
        0.75 & \multirow{2}{*}{$\vdots$} & \multirow{2}{*}{$\vdots$} & 88.1 & 30.9 & \textbf{95.3} & \textbf{53.2} \\
        1.00 & & & 89.6 & 32.0 & \textbf{95.0} & \textbf{51.4} \\
        \hline
    \end{tabular}
    \label{tab:refinement_exp_2}
\end{table}

\noindent
{\bf Results: }
Table~\ref{tab:refinement_exp_1} lists the mean cosine similarity between the generated labels for all three approaches, and the reference labels (using the Phrase-BERT embeddings), whereas in Table ~\ref{tab:refinement_exp_2}, the
percentage of potential matches, where a match is considered as a label pair receiving a cosine similarity $\ge \tau$.
The results demonstrate that across both evaluations, the refined approach consistently generates superior relationship labels than either the LLM-only method, or the rule-based RuBREx method. However, changes in \emph{temperature} have only a modest impact on the overall results, with a slight decrease in the quality of the labels as the temperature is increased.  This contradicts the original hypothesis that increasing creativity would improve \emph{``...the LLM's ability to capture the abstract nature of the relationships...''}.

The results in Table~\ref{tab:refinement_exp_2} suggest that 95\% of the labels generated by the refined label approach are similar to those in the domain-expert generated reference set (assuming a $\ge 0.6$ threshold), and in the best case (when assuming a $\ge 0.85$ threshold), up to 57.6\% of the generated labels could be considered as very similar to those in the reference set. This result suggests that by employing the ReLRAE pipeline to enrich an XML schema can significantly decrease the amount of work required by the ontology engineer and domain expert to inject additional domain knowledge into the ontology.

Another notable result is that whilst the LLM-only approach generates labels with a similar mean similarity score to those generated by RuBREx (Table \ref{tab:refinement_exp_1}), more labels were generated that were \emph{loosely} similar to the reference set, yet fewer were generated that were \emph{strictly} similar (Table \ref{tab:refinement_exp_2}).
This could be explained by the fact that RuBREx generates labels using a simple policy (represented through the rules) derived from other ontology engineering processes~\cite{bohring2005mapping} that could be similar to policy used by a domain expert when generating relationship labels.  If this is the case, then this would result in 
several cases where RuBREx generated labels are similar to those in the reference set, yet for others, the lack of context could result in labels with very little similarity.  However, although the LLM$_{\mathcal{R}}$ would not be expected to adopt such a policy to that of the domain expert,  the inclusion of contextual data regarding the relationships in the prompt for the LLM-only approach should result in labels that are similar to those in the reference set.  Thus, the superior performance observed by the Refined approach used in RELRaE could be explained due to it exploiting the synergy of these two separate (rule-based and LLM-based) approaches.

\subsection{Label Evaluation with LLM$_{\mathcal{E}}$}\label{sec:exp_eval}

One of the primary goals of the RELRaE framework is to reduce the level of time and effort required by the ontology creation process. In the Label Refinement stage RELRaE utilises one LLM to inject domain knowledge into the generation of a relationship label. Here we assess to what extent these labels are deemed acceptable, ideally by an expert. As the number of labels generated might be significantly large, we investigate whether an LLM (denoted as LLM$_{\mathcal{E}}$, to differentiate it from the LLM$_{\mathcal{R}}$ used in the refinement stage) can act as a proxy or replacement for a domain expert in assessing a label's suitability (given its associated pair of concepts), thereby addressing RQ2.

As a small number of studies have already explored the notion of using LLMs as judges or evaluators~\cite{li2024llms}; the aim of evaluation conducted here is to compare the performance of an LLM$_{\mathcal{E}}$ (in this scenario, Google's Gemini-2.0-flash model was chosen for LLM$_{\mathcal{E}}$) to that of a domain expert in assessing a given relationship label's acceptability in the context of analytical chemistry.

We initially assessed the accuracy of LLM$_{\mathcal{E}}$ as an evaluator by asking the domain expert to rate as \emph{true} or \emph{false} the score given by LLM$_{\mathcal{E}}$ to the relationship labels in the ground truth (note that this expert was different to the ones generating the labels in the reference set) using the \emph{Evaluation Prompt} (Section \ref{sec:prompt_2}). LLM$_{\mathcal{E}}$ scores the confident with which it generates the relationships using the Likert scale in Table~\ref{tab:enum}. If we consider acceptable relationships label those whose LLM confidence is \emph{Yes} or \emph{Likely}, then the domain expert's percentage of agreement with the LLM is 41.7\%. This percentage increases to 93.3\% if acceptable relationships labels are those whose confidence score assigned by LLM$_{\mathcal{E}}$ is \emph{Yes}, \emph{Likely} or \emph{Possible}.
Having gauged a sense of the accuracy of LLM$_{\mathcal{E}}$, we can now assess each of the approaches assessed in the previous evaluation (Section \ref{sec:exp_ref}). Given that the number of labels generated was very large, we assessed a random selection of 60 labels generated by each approach using LLM$_{\mathcal{E}}$ and the domain expert.

\noindent
{\bf Results: }
%
Table \ref{tab:eval_exp_2} shows that if the LLM$_{\mathcal{E}}$-based assessment is either \emph{Yes}, \emph{Likely}, or \emph{Possible} the accuracy is over  88\% for all approaches. Notably, the approach with the best performance is the Refined one, where LLM$_{\mathcal{E}}$-based can confidently assign a correct relationship label, even though the confidence is moderate (\emph{Possible}). This seems to suggest that the effect of having rule generated labels is beneficial.


\begin{table}[t]
    \centering
    \setlength{\tabcolsep}{10pt} 
    \caption{Percentage of relationship labels found acceptable, either based on a tighter acceptability criteria on the Likert scale - ``Likely'' (4) or ``Yes'' (5), and with a looser criteria - ``Possible'' (3) , ``Likely'' (4), or ``Yes'' (5).}
    \begin{tabular}{l | c c}
        & Likert Score 4,5 (\%) & Likert Score 3,4,5 (\%) \\
         \hline
         RuBREx & 33.3 & 88.3 \\
         LLM-only & 36.7 & 91.7 \\
         Refined & 43.3 & 95.0 \\
         \hline
    \end{tabular}
    \label{tab:eval_exp_2}
\end{table}